\documentclass[letterpaper, 10 pt, conference]{ieeeconf}  % Comment this line out if you need a4paper

\IEEEoverridecommandlockouts                              % This command is only needed if 
\usepackage{graphics} % for pdf, bitmapped graphics file
\usepackage{epsfig} % for postscript graphics files
\usepackage{subfigure}
\usepackage{amsmath} % assumes amsmath package installed
\usepackage{amssymb}
\usepackage{amsfonts} 
\usepackage{booktabs}% http://ctan.org/pkg/booktabs

\usepackage{todonotes} % for comments

\title{\LARGE \bf
Neural Ordinary Differential Equations for Nonlinear System Identification
}

\author{Aowabin Rahman, J\'an Drgo\v na, Aaron Tuor, and Jan Strube % <-this % stops a space
\thanks{Funding for this work was provided by the Pacific Northwest National
Laboratory’s (PNNL) Laboratory Directed Research and Development (LDRD) Program.
PNNL is a multiprogram national laboratory operated by Battelle for the United States Department of Energy under DEAC05-
76RL01830.}% <-this % stops a space
\thanks{The authors are with the Pacific Northwest National Laboratory, Richland, Washington, USA
         \{aowabin.rahman, jan.drgona, aaron.tuor, jan.strube\}@pnnl.gov }%
}

\begin{document}

\maketitle
\thispagestyle{empty}
\pagestyle{empty}

%%%%%%%%%%%%%%%%%%%%%%%%%%%%%%%%%%%%%%%%%%%%%%%%%%%%%%%%%%%%%%%%%%%%%%%%%%%%%%%%
\begin{abstract}

Neural ordinary differential equations (NODE) have been recently 
proposed as a promising approach for nonlinear system identification tasks. 
In this work, we systematically compare their predictive performance with current state-of-the-art nonlinear and classical linear methods. In particular, we present a quantitative study comparing NODE's performance against neural state-space models and classical linear system identification methods. We evaluate the inference speed and prediction performance of each method on open-loop errors across eight different dynamical systems. 
The experiments show that NODEs can consistently improve the prediction accuracy by an order of magnitude compared to benchmark methods.
Besides improved accuracy, we also observed that NODEs are less sensitive to hyperparameters compared to neural state-space models.  On the other hand, these performance gains come with a slight increase of computation at the inference time.
\end{abstract}

%%%%%%%%%%%%%%%%%%%%%%%%%%%%%%%%%%%%%%%%%%%%%%%%%%%%%%%%%%%%%%%%%%%%%%%%%%%%%%%%
\section{Introduction}
\label{sec:intro}

Nonlinearity is a key characteristic of dynamical systems observed in many scientific and engineering domains \cite{Khalil:1173048}. Traditionally, numerical methods to solve differential equations are used to describe the time evolution of system observations; however, these methods often require expert knowledge and/or knowledge of underlying governing equations, and so can be restrictive when applied to system identification of unknown systems. As such, data-driven models are often feasible alternatives for system identification.

Historically, dicrete-time, time-invaraint, linear models have been considered for system identification tasks \cite{van2012subspace}. However, recent research has made significant strides in leveraging machine (ML)/deep learning (DL) approaches for solving differential equations \cite{battaglia2016interaction, jia2019neural, rackauckas2019diffeqflux, innes2019differentiable, baker2019workshop}, which has made nonlinear ML models attractive candidates for system identification tasks. Approaches such as physics-informed neural networks (PINNs) are capable of solving partial differential equations (PDEs) \cite{lu2021deepxde}, but require knowledge of underlying equations. Deep learning models have also been developed as black-box models for system identification of unknown dynamical systems \cite{ogunmolu2016nonlinear, krishnan2017structured}. Neural state-space models (NSSMs)~\cite{SUYKENS95,NIPS2018_8004,MastiCDC2018} have been shown to be flexible to allow various degrees of domain knowledge to be embedded in the model formulation \cite{skomski2021automating,skomski2021constrained,tuor2020constrained}. However, one potential limitation of such models may be that they could be subject to errors due to discrete-time formulation.

 Neural Ordinary Differential Equations (NODEs) have recently emerged as a new class of ``continuous time/depth" deep learning models \cite{Chen2018,massaroli2020dissecting}. Instead of  stacking multiple discrete hidden layers end-to-end, NODEs are constructed by defining the derivative using a neural network. Subsequently, the outputs of the NODE models are computed by using an off-the-shelf ODE solver, thus generating a continuous output. Recently NODEs are beginning to be deployed in system identification tasks \cite{quaglino2020snode, alvarez2020, zhong2019symplectic, DBLP:journals/corr/abs-2102-06794}. 
 The NODE formulation can be particularly suitable for system identification tasks, as real-world phenomena are mostly continuous time-evolution processes (which are often unknown) that generate continuous observations. Therefore, a predictive model to forecast system behavior should consider temporal integration methods that are typically used to solve ODEs \cite{alvarez2020}.

The  contribution of this paper is to benchmark the performance of
NODEs in capturing the nonlinear behavior in eight different nonlinear dynamical systems. The performance is benchmarked against traditional linear subspace models and neural state-space models (NSSMs). 
The eight considered systems in the numerical experiments include both autonomous and non-autonomous systems, as well as chaotic and non-chaotic systems. 
We assess the open-loop prediction accuracy, sensitivity to hyperparameter selection, and inference time. We demonstrate that NODEs can achieve a magnitude higher accuracy than NSSMs, while being less sensitive to the choice of hyperparameters. However, these potential benefits may come at an expense in inference time with roughly an order of magnitude slower model. 
\section{Methods}
\label{sec:methods}
We consider a non-autonomous, partially observable nonlinear dynamical system with latent states $\mathbf{x}(t) \in \mathbb{R}^{n_x}$, observations $\mathbf{y}(t) \in {\mathbb{R}}^{n_y}$ and control inputs/disturbances $\mathbf{u}(t) \in \mathbb{R}^{n_u}$, which can be described as follows: 
\begin{align}
    \frac{d \mathbf{x(t)}}{dt} &= f(\mathbf{x}(t), \mathbf{u}(t))\label{diffeq_1a} \\
    \mathbf{y}(t) &= g(\mathbf{x(t)}) \label{diffeq_1b} \\
    \mathbf{x}(t_0) &= \mathbf{x_0} 
\end{align} 

For learning the models we assume having a dataset $D$ of system measurements, corresponding to the input-output tuples of observed trajectories with sampling time $\Delta$. 
\begin{equation}
    \label{eq:dataset}
    D = \{(\mathbf{u}^{(i)}_t, \mathbf{y}^{(i)}_t), (\mathbf{u}^{(i)}_{t+\Delta}, \mathbf{y}^{(i)}_{t+\Delta}),  \ldots, (\mathbf{u}^{(i)}_{t+N\Delta}, \mathbf{y}^{(i)}_{t+N\Delta})\},
\end{equation}
here $i = 1,2,\ldots, n$ represents $n$ different batches of the system input-output trajectories in the length of $N$ steps.

\subsection{Neural Ordinary Differential Equation (NODE) Model}

Previous works \cite{haber2017stable, chashchin2019predicting} have shown that  deep neural networks with skip connections can be interpreted as Euler discretizations of an underlying dynamical system. As the step size approaches zero, the time-evolution of the hidden units can be represented as a black-box neural differential equation~\eqref{eq:dh_t}, that can be solved using an ODE solver \cite{chen2018neural}.
\begin{equation}
    \frac{d \mathbf{x}}{dt} = \mathbf{f}(\mathbf{x}(t), t,\mathbf{\theta})
\label{eq:dh_t}
\end{equation}
Here, $\mathbf{f}(\mathbf{x(t)}, t,\theta)$ refers to the vector field that is parmeterized by a deep neural network forming a neural ordinary differential equation (NODE). The time evolution of the states $\mathbf{x}(t)$ can therefore be expressed as time integration of the NODE system~\eqref{eq:dh_t} using  an off-the-shelf ODE solver:
\begin{align*}
    \mathbf{x}(t_1) &= \mathbf{x}(t_0) + \int_{t_0}^{t_1}   \mathbf{f}(\mathbf{x}(t), t, \mathbf{\theta}) dt \nonumber\\
                    &= ODESolve(\mathbf{x}(t_0), \mathbf{f}, t_0, t_1, \mathbf{\theta}) 
\end{align*}
where $t$ represents time, $\mathbf{x}(t)$ represent the latent space and parameters $\theta$ represent weights and biases of the neural network $\mathbf{f}$.
Contrary to classical neural network optimization approaches leveraging reverse mode automatic differentiation (backpropagation), the parameters of NODE~\eqref{eq:dh_t}  are optimized using the adjoint sensitivity method \cite{pontryagin1987mathematical, chen2018neural}.

In this paper, we use a NODE system proposed by Massaroli et al. \cite{massaroli2020dissecting} called data-controlled NODE given as follows:
\begin{equation}
    \frac{d \mathbf{x}}{dt} = \mathbf{f}(\mathbf{x}(t), t, \mathbf{x}(t_0), \theta)
\label{eq:dh_t_DC}
\end{equation}
Thus, in the NODE with data control, the vector field is a function of input data -- which in our case is $\mathbf{x}(t_0)$. This acts as a regularizer, constraining the solution to be smooth with respect to $\mathbf{x_0}$ \cite{massaroli2020dissecting}. The initial condition in a NODE with data-control is:
\begin{equation}
    \mathbf{x}(t_0) = g_x(\mathbf{y}_0, \mathbf{u}(t_0))
\end{equation}
Thus, the latent space at time $t_0$, $\mathbf{x}(t_0)$ is modeled as a function of the observations $\mathbf{y_0}$ and control inputs $\mathbf{u}(t_0)$. To deal with the partially observable systems, we introduce a mapping of latent space to the observable space as follows:
\begin{equation}
    \mathbf{y}(t) = g_y(\mathbf{x}(t))
\end{equation}

Here, $g_x$, is a non-linear maps parametrized by a neural network, and $g_y$ is a linear map.

\subsection{Linear Subspace Identification Methods}

As a baseline for the comparison, we consider classical linear subspace methods to identify linear state space models (LSSM) given as follows:
\begin{align}
    \mathbf{x}_{t+1} &= \mathbf{A}\mathbf{x}_t + \mathbf{B} \mathbf{u}_t +  \mathbf{K}\mathbf{y}_{t}\\
    \mathbf{\hat{y}}_{t+1} &= \mathbf{C} \mathbf{x}_{t+1}
\end{align} 
Here $\mathbf{A}$, $\mathbf{B}$, and $\mathbf{C}$ are identified system matrices, while $\mathbf{K}$ represents Kalman filter gain.
Linear subspace models have historically been popular for system identification tasks due to their relative simplicity and computational speed, with applications dating back 30 years \cite{van2012subspace}. These models use linear functions as both state estimator (i.e. mappping observations to a latent space) and state space models \cite{favoreel2000subspace}. In this paper, we consider the following three methods: Numerical Algorithms for Subspace State Space System Identification (N4SID) \cite{van1994n4sid}, Multi-Variable Output-Error State Space (MOESP) \cite{van1995unifying}, and Canonical Variate Analysis (CVA) \cite{van2012subspace}. The selection of the specific method among the three used is considered as a hyperparameter.

\subsection{Neural state-space models}

More recently, neural state-space models (NSSM) have been applied   to perform well on the identification tasks across multiple dynamical systems~\cite{LatentDynamics2018,MastiCDC2018,NIPS2018_8004}.
In this paper we consider the following definition of the NSSM~ \cite{skomski2021constrained}:
\begin{align}
  \mathbf{x_0} &= \mathbf{f_{o}}(\mathbf{y}_{1-N_p}; ....;\mathbf{y}_0) \\
    \mathbf{x}_{t+1} &= \mathbf{f_x}(\mathbf{x}_t) + \mathbf{f_u} (\mathbf{u}_t) \label{eq:fxfu}\\
    \mathbf{\hat{y}}_{t+1} &= \mathbf{f_y} (\mathbf{x}_{t+1}) \label{eq:fy}
\end{align} 
Here, $\mathbf{f_0}$ , $\mathbf{f_x}$ and $\mathbf{f_u}$ are non-linear functions parametrized by deep neural networks, whereas $\mathbf{f_y}$ is a linear map. We can embed prior information in $\mathbf{f_x}$ and/or $\mathbf{f_y}$ through the linear maps associated with the models.  In this study, we have considered the ``soft-SVD" regularization technique introduced in  \cite{skomski2021constrained}, in which the linear maps in the linear/non-linear models are factorized using singular value decomposition (SVD). The block NSSM model also allows constraints on outputs using a penalty method \cite{tuor2020constrained}.

\section{Case Studies}
\label{sec:case_studies}

\subsection{Description of nonlinear systems}
\label{sec:nonlinear_sys}

We evaluate the performance of the NODE, neural SSM, and linear SSM models on eight dynamical systems.

\paragraph{Aerodynamic body (aero)} We consider a modeling task of a non-autonomous airplane model. The dataset contains 501 samples (i.e. timesteps), with 10 inputs and 5 outputs. The inputs consist of angles of control surfaces, angle of attack and sideslip and the angular velocities along x, y and z. The data for this system was obtained from \cite{aero2021}.
    
    \paragraph{Continuous Stirred Tank Reactor (CSTR)} This is a mathematical model of a chemical reactor representing non-autonomous, non-dissipative system exhibiting oscillatory behavior. The two outputs of interest are the concentration of the products and reactor temperature respectively, which are computed using a set of ODEs. In this paper, we used an emulator~\cite{PSL_lib} using the \textit{SciPy} ODEInt package to generate a dataset of 12000 timesteps.

    \paragraph{Double Pendulum} Here we consider an autonomous, chaotic, highly oscillatory system. The four outputs of interest are the angular positions and velocities of the two pendulums. Using the governing ODEs \cite{doublependulum2021}, we use the above-mentioned emulator \textit{SciPy} ODEInt to generate a dataset of 2000 timesteps.
    
    % \paragraph{Building} This is a real-world  dataset representing zone temperatures in a commercial building, which can be modeled as a non-autonomous system \cite{drgovna2021physics}. There are 20 outputs in the dataset corresponding to temperatures in 20 thermal zones, 40 inputs corresponding to supply temperatures and mass-flow rates of the HVAC system, and one disturbance corresponding to ambient temperatures \cite{drgovna2021physics, rubio2017learning}. The dataset consists of 12,551 timesteps sampled at 15-min intervals. 
    
   \paragraph{Vehicle} This is a mathematical model of a vehicle non-autonomous dynamics system \cite{vehicle2021}. The three outputs of interest are  the yaw-rate and the longitudinal and the lateral velocities. The five inputs are the slip on the four tires and the steering angle. The dataset contains 2501 timesteps.
    
   \paragraph{Tank} This is a single-input-single-output (SISO) model of a non-autonomous system \cite{tank2021}. The input consists of a voltage to a pump that controls the inflow of fluid to an upper tank. The upper tank has an opening at the bottom allowing fluid flow to a lower tank. The target output is the fluid level in the lower tank. The tank dataset consists of 3000 timesteps.
    
    \paragraph{Two-tank} Similar to the \textit{tank} system, the \textit{two-tank} is also a non-autonomous system
    also consisting of two tanks connected by a valve. The key difference between the two systems is that \textit{two-tank} has two control variables: the pump speed and the opening of the valves between the pumps. The two outputs are the fluid levels in the two tanks. Using the emulator (developed using \textit{SciPy} ODEInt) based on governing ODEs implemented in \cite{PSL_lib}. The dataset contains 12000 timesteps.

   \paragraph{Pendulum} This is an autonomous system describing the simulated behavior of an ideal pendulum \cite{schmidt2009distilling}. The dataset consists of 493 timesteps and the outputs are angular position, velocity and acceleration.
    
    \paragraph{Linear Oscillator} This is a dataset describing the behavior of an actual two-mass spring system obtained from \cite{schmidt2009distilling}, which can be modeled as an autonomous system. The dataset consists of 870 timesteps and the target outputs are linear position, velocity and acceleration.

\subsection{Details of numerical experiments}
\label{sec:num_expt}

The numerical experiments were performed by equally splitting each dataset (corresponding to each of the eight systems described above) into a training, development and test set. 
We performed a grid search to select the hyperparameters for each of the considered methods.

\paragraph{Linear state space models}
For the LSSM, we consider the following hyperparameters:
\begin{itemize}
    \item Method $\in$ \{N4SID, MOESP, CVA\}
    \item Size of latent space $\in \{10, 20, 40, 60, 80 \}$
    \item Forecast horizon $\in \{1, 5, 10, 20, 50 \}$
\end{itemize}
We use the Systems Identification Package for PYthon (SIPPY) \cite{Sippy2021} to train our linear subspace models. We evaluate the performance of all three models using the open-loop mean squared error (MSE) metric.

\paragraph{Neural state space models}
The hyperparameters for the NSSM, and the corresponding search space were:
\begin{itemize}
    \item Linear map  $\in$ \{Linear, Soft-SVD\}
    \item Non-linear model for $\mathbf{f_x}, \mathbf{f_u}$ $\in$ \{Linear, MLP, RNN\}
    \item Loss coefficient $Q_{dx}$ $\in$ \{0, 0.1, 0.2 \}
    \item Forecast horizon $N_{steps}$ $\in$  \{1, 5, 10, 20, 50\}
    \item Latent space multiplier $N_x$ $\in$  \{10, 30, 50\}
\end{itemize}
To constrain the hyperparamter search for NSSM, we assume that both $\mathbf{f_x}$ and $\mathbf{f_u}$ have an identical model architecture. We also assume that the number of past observations, $N_p$ is equal to the prediction horizon, $N_{steps}$. The loss coefficient $Q_{dx}$ weights the state smoothing term used to regularize the SSM models, relative to the reference loss.  We assume that the size of the latent space ($\mathbf{x}$), $n_x$ =  $N_x$ $\times$ $n_y$, where $n_x$ is a hyperparameter. 
We implement the NSSM using 
NeuroMANCER~\cite{Neuromancer2021} library based on Pytorch~\cite{paszke2019pytorch} and train the models using the AdamW optimizer \cite{adamW2017} with a learning rate of 0.003 for 5000 epochs. 

When training the NSSM  models, we employed downsampling specifically for two of the eight datasets: \textit{tank} and \textit{vehicle} datasets. We resampled these two datasets at a frequency of 10 and 8 datapoints respectively. We used downsampling on these two datasets for the NSSM models, as the transients associated with these two time series have a longer timescale. When training the NODE model we did not use any downsampling.

\paragraph{Neural ordinary differential equations}
For our NODE model there were fewer hyperparameters to select compared to the SSM, as presented below:
\begin{itemize}
    \item Latent space multiplier $\in$ \{1, 5, 10\}
    \item Size of hidden space for NODE $\in$ \{32, 64, 128, 256\}
    \item Size of hidden layer of NODE $\in$ \{32, 64, 128, 256\}
\end{itemize}
We model the vector field using a multi-layered perceptron (MLP) with one hidden layer, where the size of the hidden layer ($\mathbf{f}$ in equation \ref{eq:dh_t_DC}) is a hyperparameter. Note that for the NODE model, we consider $N_p = N_{steps} = 1$, as we found that considering an n-step loss did not significantly improve the open-loop predictions. 
The NODE models are implemented in Python using the \textit{torchdyn} package \cite{poli2020torchdyn}. 
We train our NODE models with a learning rate of 0.01 using the ADAM optimizer \cite{kingma2014adam}. We use the Dormand-Prince method \cite{kimura2009dormand} (which uses an adaptive step size) as our ODE solver.

\section{Results and Analysis}
\label{sec:results}

\begin{figure*}[thbp!]
    \subfigure[]{\includegraphics[scale=0.2]{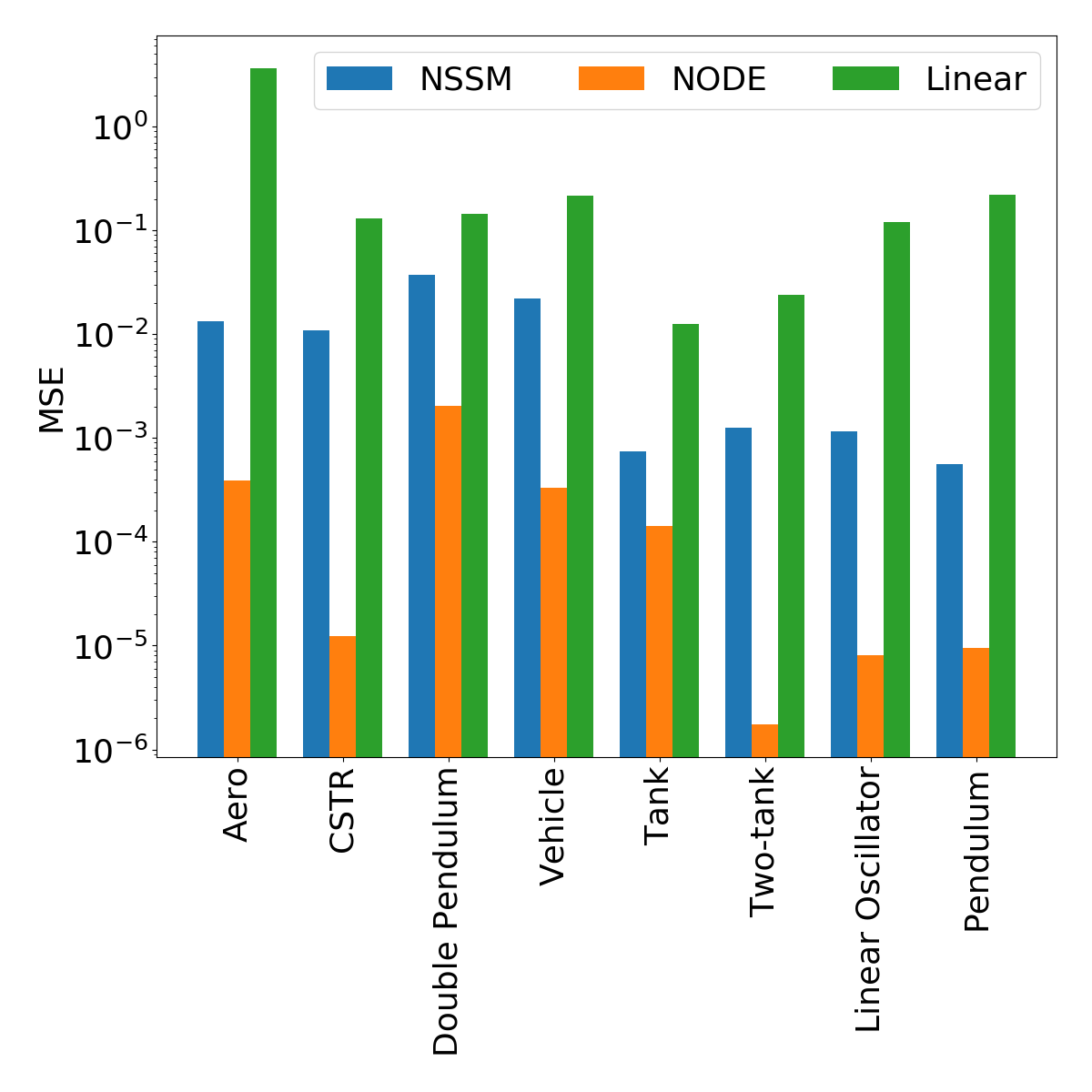}}
    \subfigure[]{\includegraphics[scale=0.2]{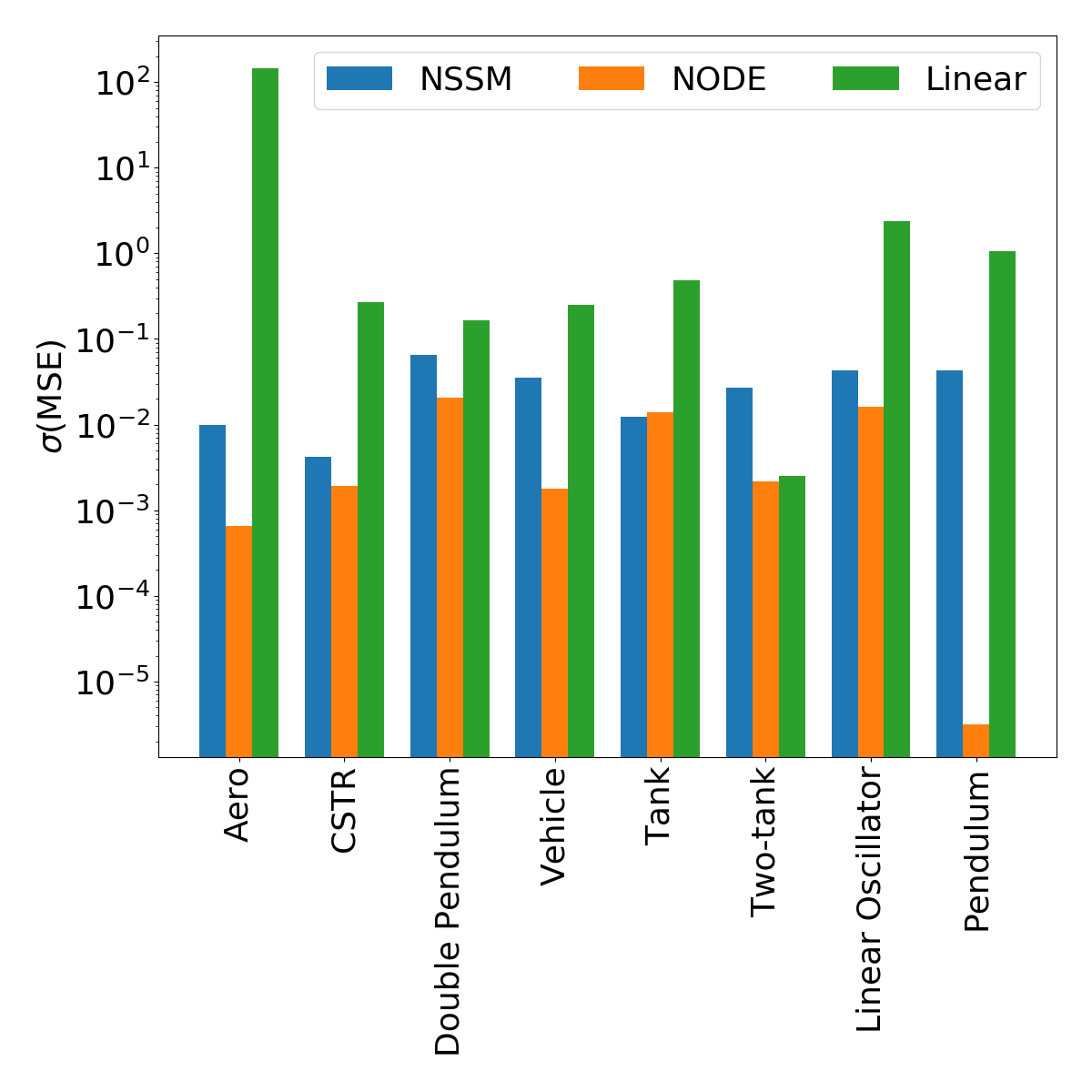}}
    \subfigure[]{\includegraphics[scale=0.2]{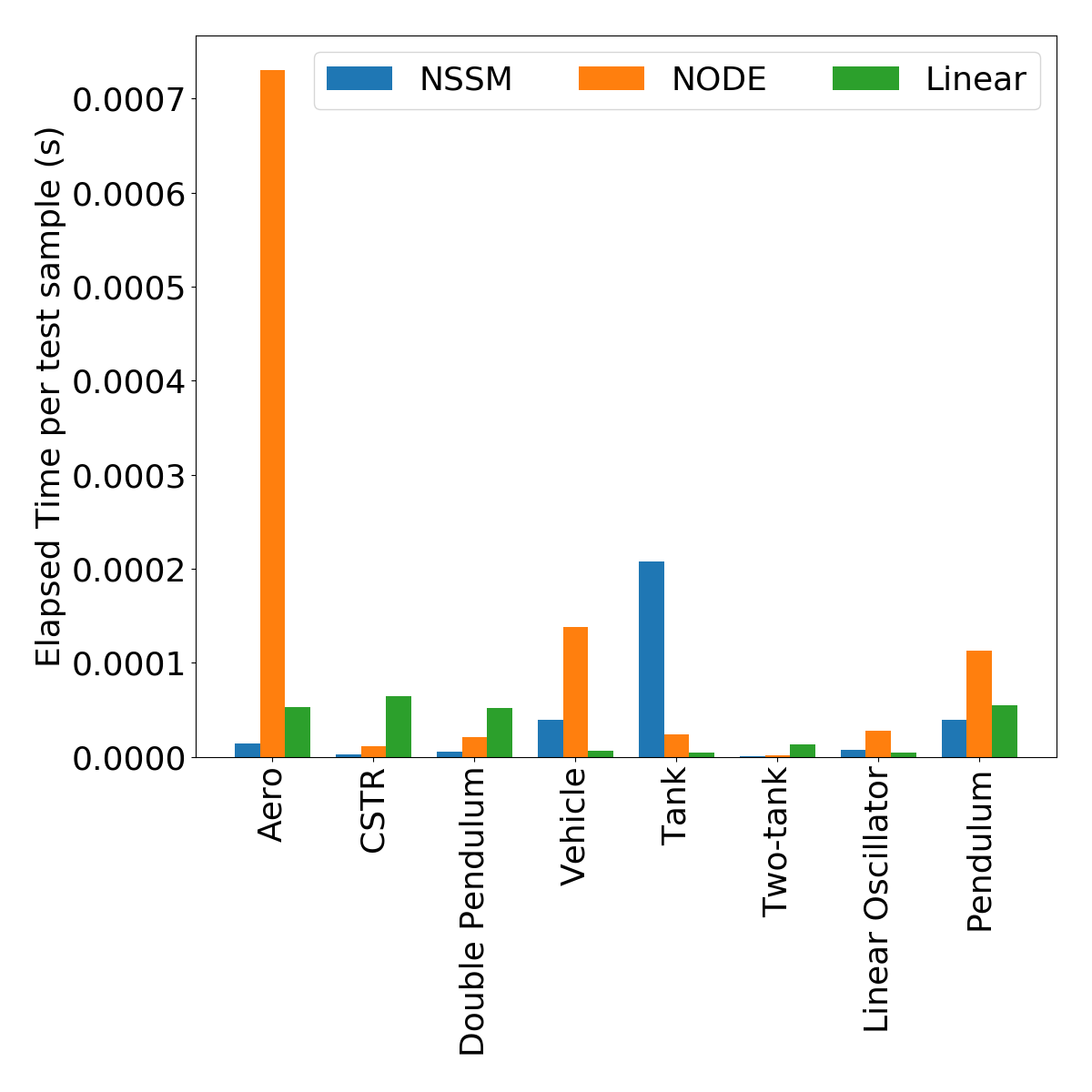}}
      \caption{Comparison of model performance: (a) Model accuracy, in terms of mean squared error (MSE) on test set; (b) Standard deviation in MSE in test set for NODE and neural SSM; (c) Inference time associated with NODE and neural SSM. Note the logarithmic scale on the y-axis for figures (a) and (b).  }
   \label{fig:bar_plots}
   \end{figure*}

\begin{figure*}[thbp!]
    \centering
    \subfigure[]{\includegraphics[width=0.45\textwidth, trim=70 23 70 60, clip]{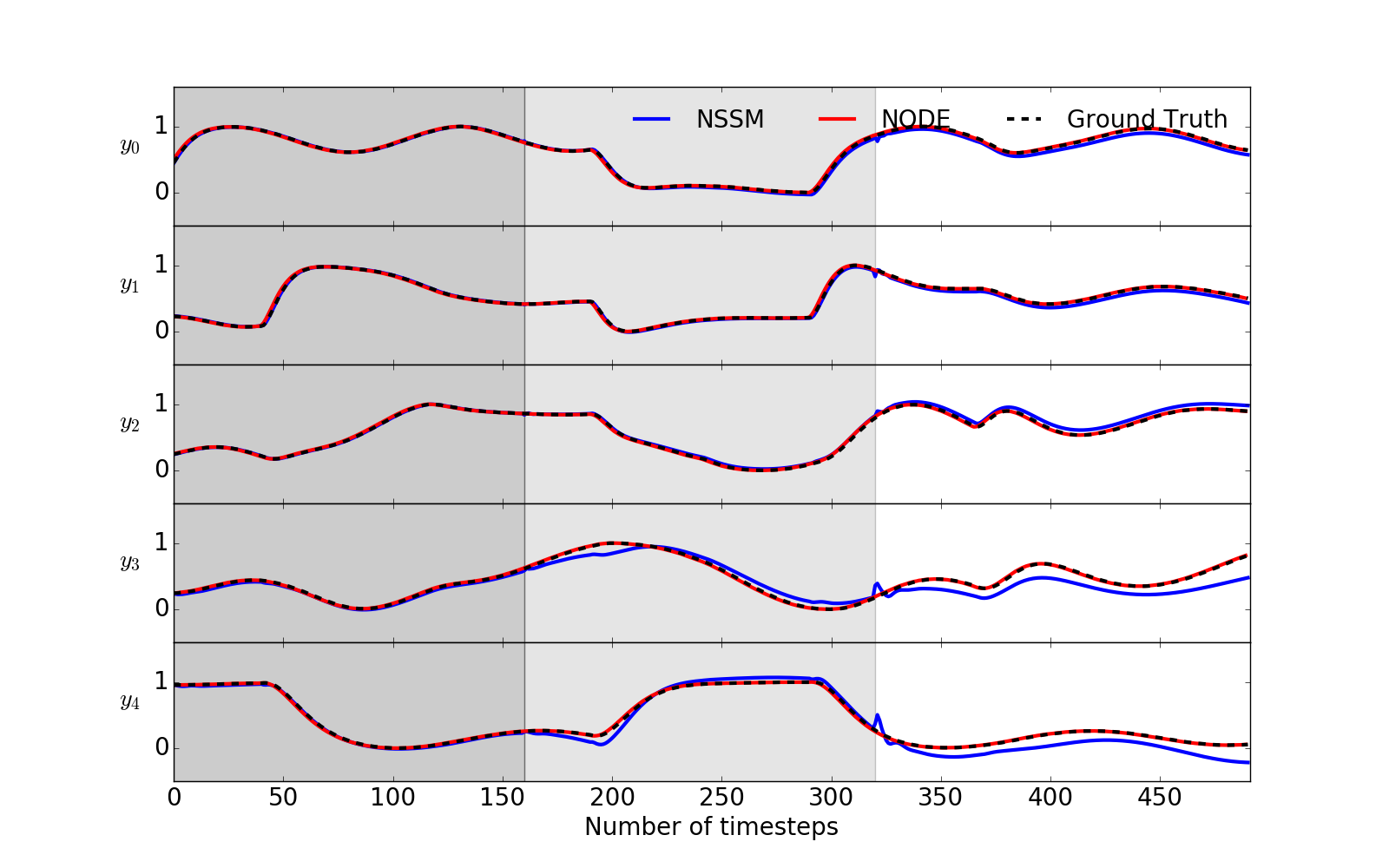}} \hfill
    \subfigure[]{\includegraphics[width=0.45\textwidth, trim=70 23 70 60 clip]{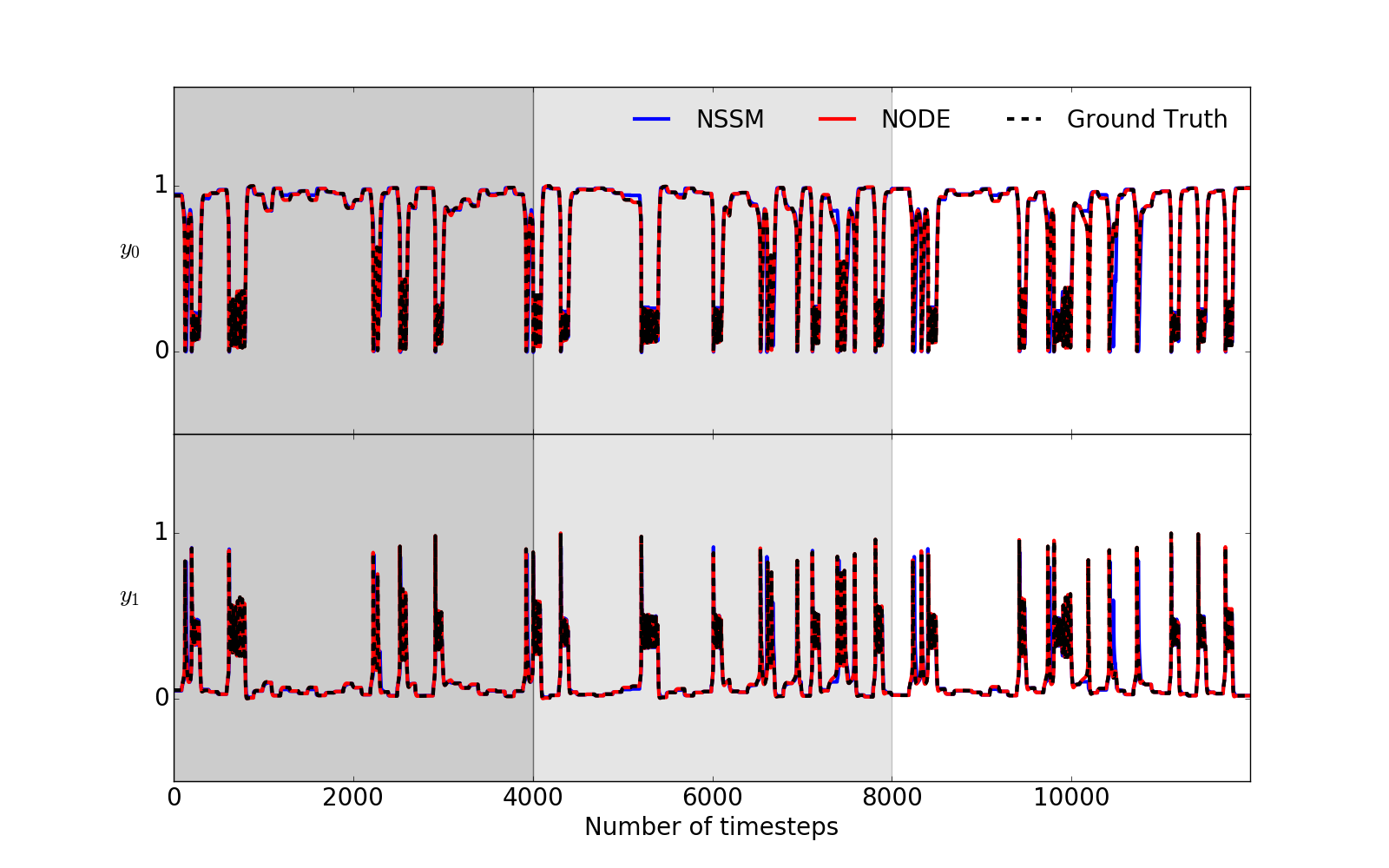}} \\
    \subfigure[]{\includegraphics[width=0.45\textwidth, trim=70 23 70 60, clip]{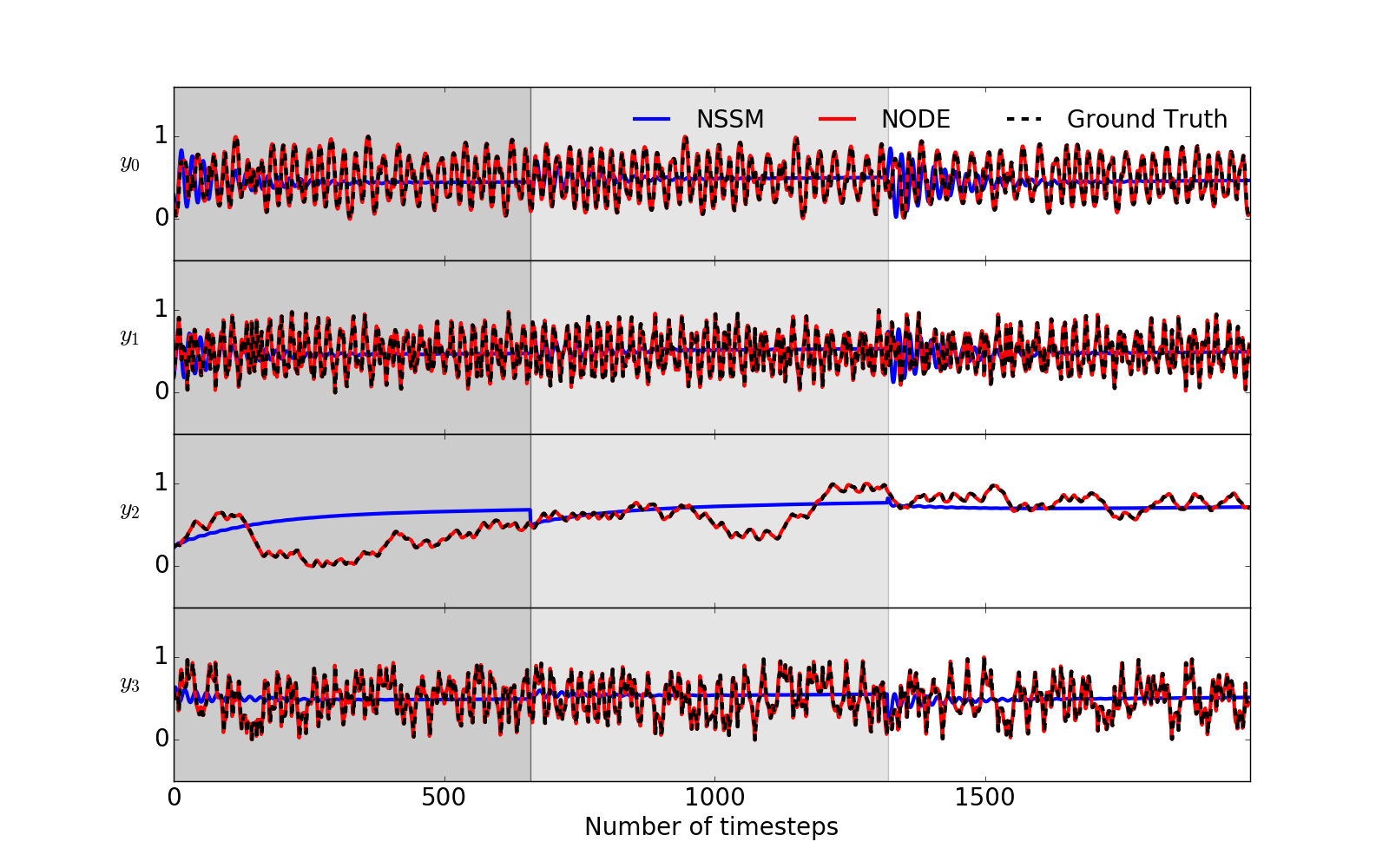}} \hfill 
    \subfigure[]{\includegraphics[width=0.45\textwidth, trim=70 23 70 60, clip]{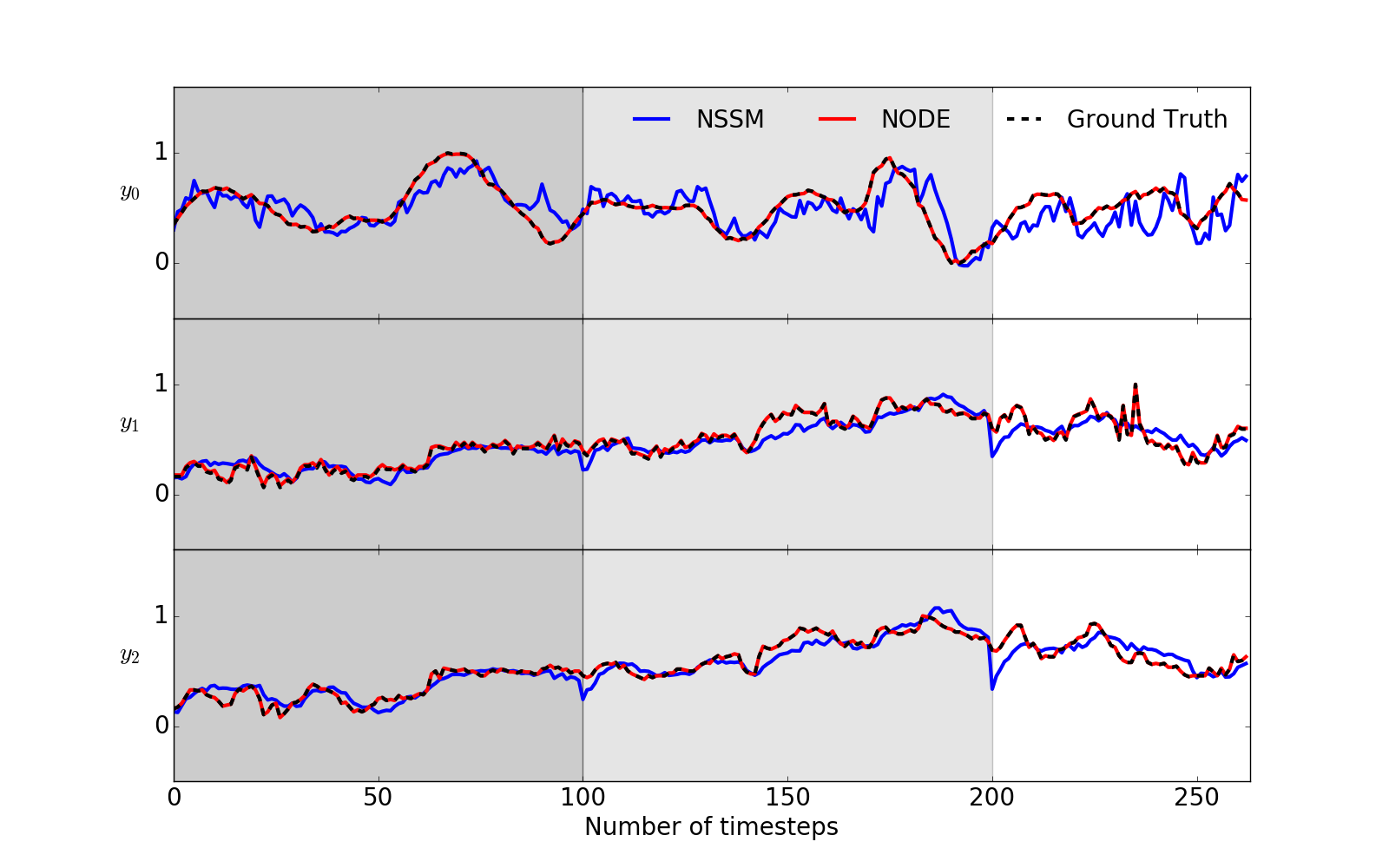}} \\
    \subfigure[]{\includegraphics[width=0.45\textwidth, trim=70 23 70 60, clip]{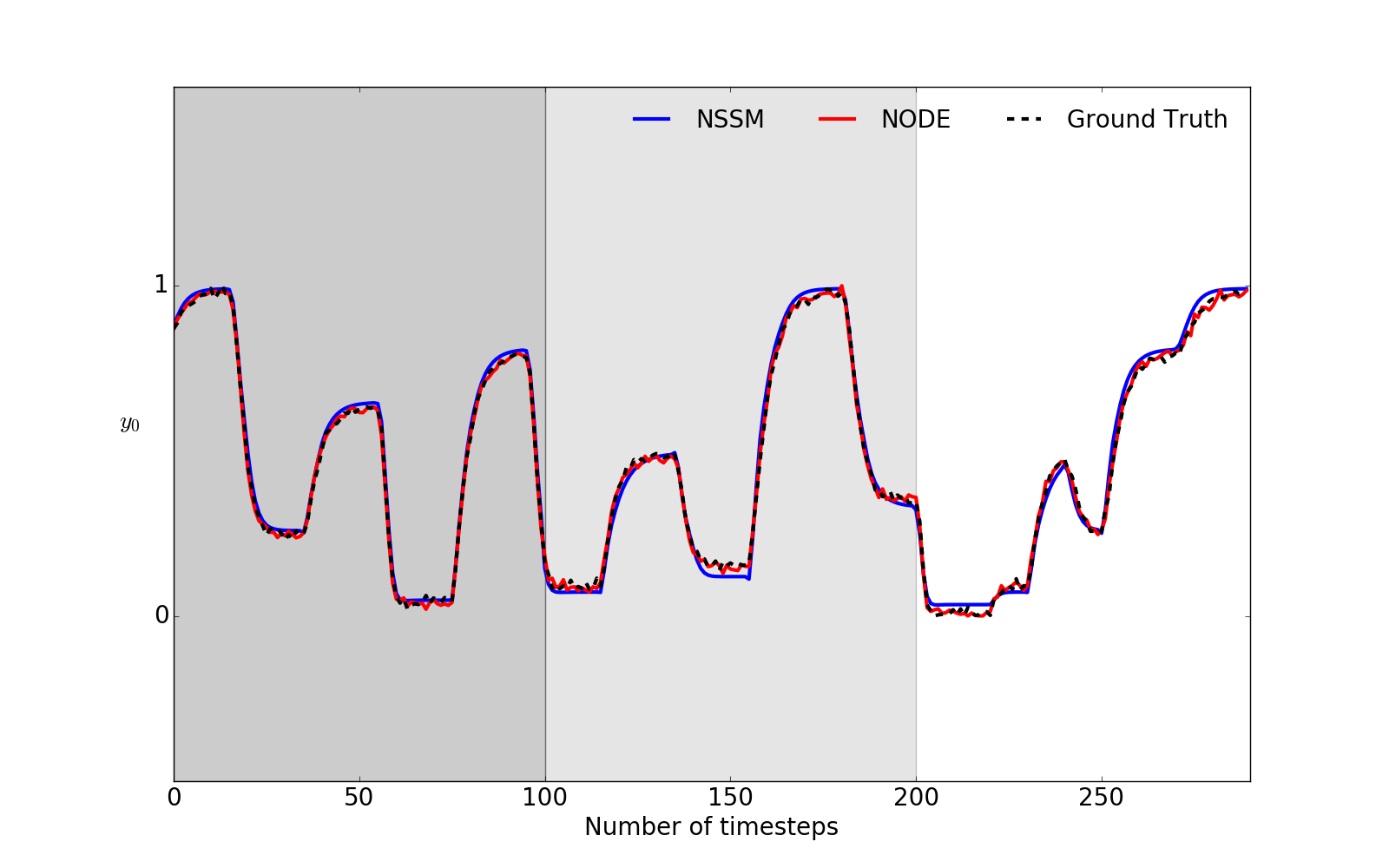}} \hfill
    \subfigure[]{\includegraphics[width=0.45\textwidth, trim=70 23 70 60, clip]{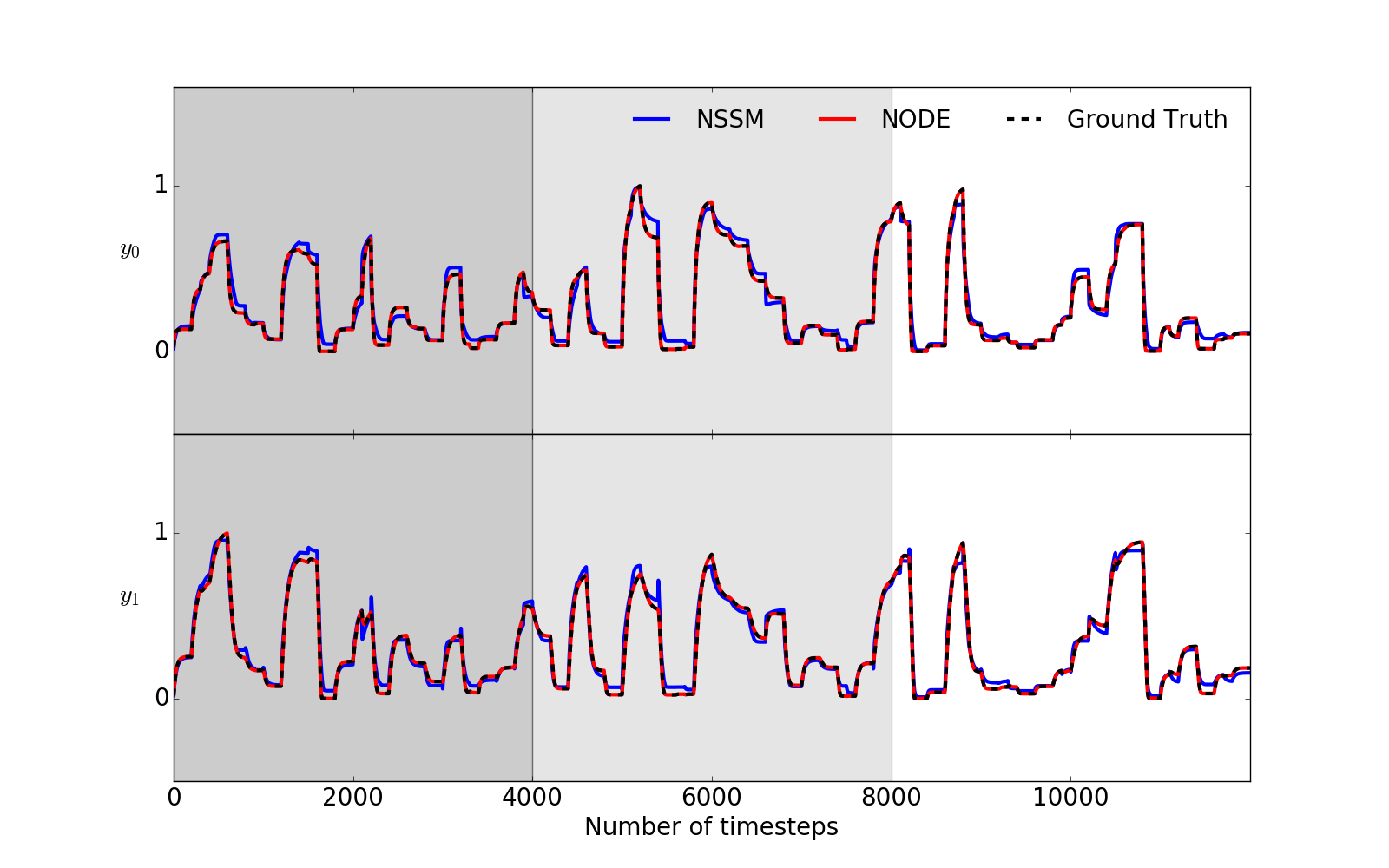}}
    \subfigure[]{\includegraphics[width=0.45\textwidth, trim=70 23 70 60, clip]{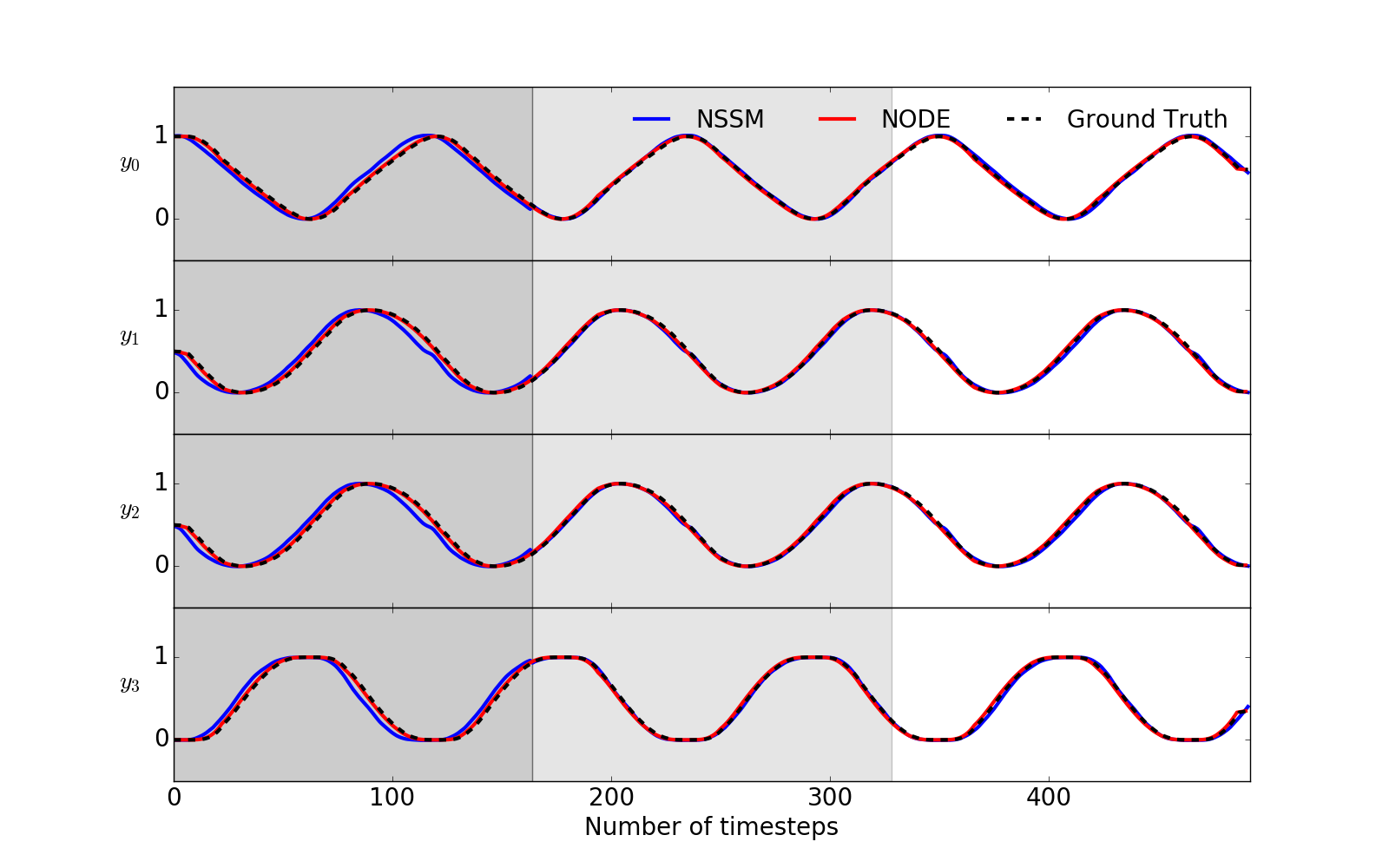}}\hfill
    \subfigure[]{\includegraphics[width=0.45\textwidth, trim=70 23 70 60, clip]{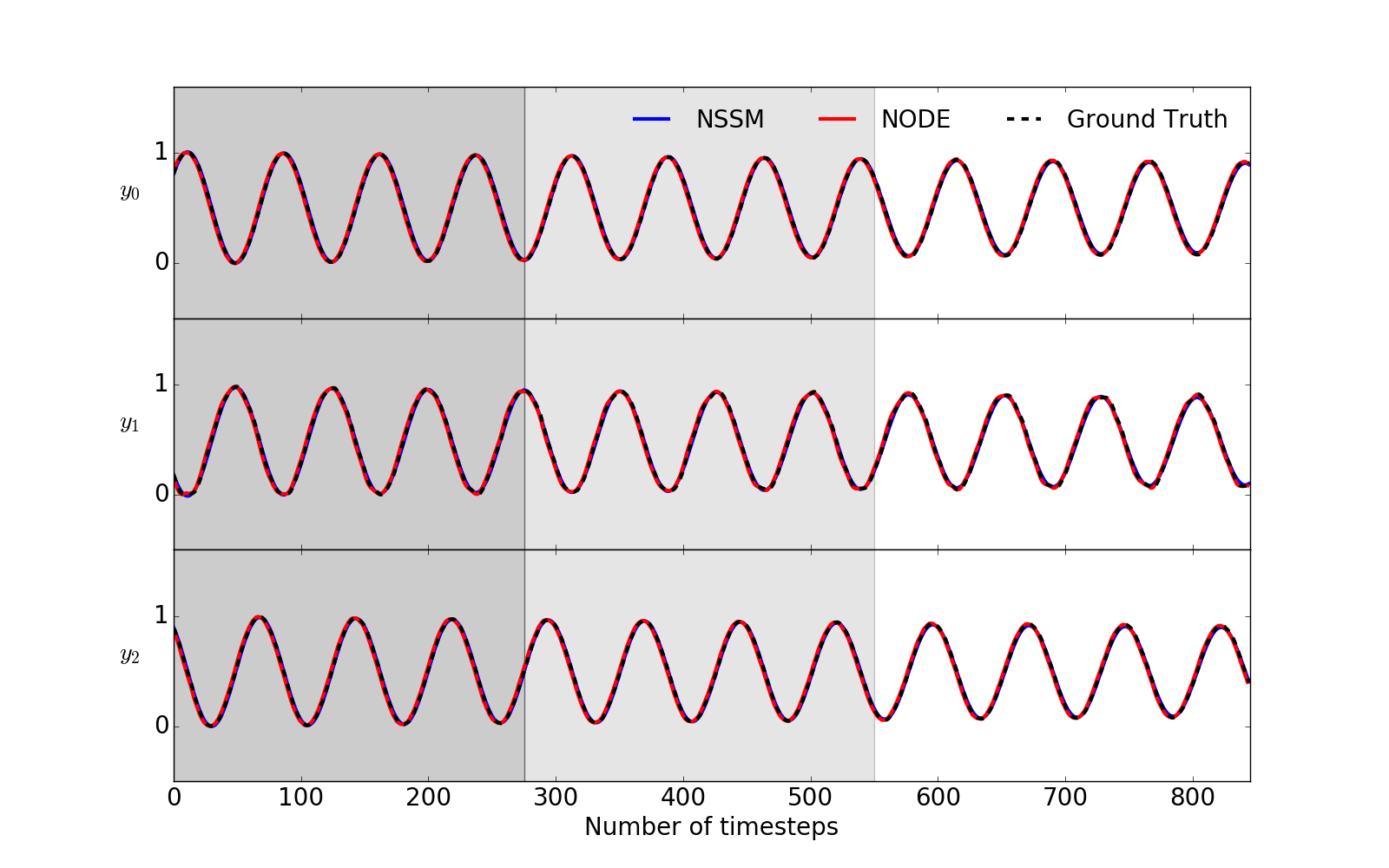}} 
      \caption{Comparison of open-loop trajectories obtained using NODE vs. those obtained using SSM for the following dynamical systems: (a) Aero; (b) CSTR; (c) Double Pendulum; (d)  Vehicle; (e) Tank; (f) Two-tank (g) Penduulum and (h) Linear oscillaator. Train, development, and test set division of the dataset is depicted by a gray shading.}
   \label{fig:trajectories}
   \end{figure*}

A comparison between the NSSM, NODE and the LSSM is presented in Fig.~\ref{fig:bar_plots}. As we may expect, we oberve that both the  NSSM and the NODE outperform the linear subspace models. We observe that the open-loop mean squared error (MSE) for the test set associated with the NODE is lower than the corresponding MSE for the NSSM for all nine systems analyzed. The reduction in MSE is by a factor of $\sim O(10^2)$ to $\sim O(10^3)$ for all nine systems. This observation highlights the potential benefits in deploying continuous time models for learning unknown dynamical systems.  We observe that the reduction in open-loop MSE is the highest for the \textit{CSTR} and the \textit{Two-tank} systems. The trajectories for these two datasets are presented in figures  \ref{fig:trajectories}(b)) and (f). Note that both datasets contain $> 10,000$ data points, which, in the context of the datasets analyzed, is in the high-data regime. A formal ablation study comparing the performance of NODE and NSSM could be the focus of future research.

A comparison of the trajectories obtained using the best-performing hyperparameter sets for NODE and NSSM is provided in Fig.~\ref{fig:trajectories}. We observe that, while NSSM can effectively learn the behavior of simple, autonomous systems such as the linear oscillator in Fig.~\ref{fig:trajectories}(h), it might struggle on autonomous systems exhibiting chaotic behavior such as the double pendulum in Fig.~\ref{fig:trajectories}(c). It is possible that NSSM could have performed well on hyperparameter combinations outside our grid search; however this in itself highlights one of the potential benefits of using NODE. 

Fig.~\ref{fig:bar_plots}(b) shows that the standard deviation of the test MSE, $\sigma (MSE)$ for NODE is approximately equal to lower than the corresponding $\sigma (MSE)$ for NSSM across nine systems analyzed. Again, the relative improvement in $\sigma (MSE)$ could be as much as three orders of magnitude, compared to the NSSM models. Thus, we find that the NODE can be deployed with a relatively few set of hyperparameters compared to NSSM, and the open-loop MSE may not be as strongly dependent on selection of these hyperparameters compared to the NSSM models. We also observed that NODE may be able to capture long-term transients without any downsampling of data, as observed from the trajectories of the \textit{vehicle} system in Fig~\ref{fig:trajectories}(d). 
It is likely that using a continuous time-based model with an adaptive step size can alleviate the need for extensive hyperparameter search and data sampling strategies.

However, the differential equation solvers in NODE can come with an increased computational cost \cite{lehtimaki2021accelerating}. Figure \ref{fig:bar_plots}(c) compares the elapsed time per timestep/sample in the test set for all three models considered. We observe that for eight out of the nine systems (all except the \textit{tank} system), the inference time per sample is greater for NODE compared to NSSM. For the aero dataset, the relative increase in the inference time comes with a factor of $\approx50$; but for the other seven systems, the relative increases are by factors of $\approx1.1$ to $4.4$. Note that the linear subspace model actually reports a higher inference time per sample for the some of the systems considered. A possible reason for this could be that in the SIPPY implementation, the inference operation is not parallelized, and so the elapsed time can strongly depend on the number of samples in the dataset.

\section{Conclusions}
\label{sec:conclusions}

In this paper, we report a systematic case study comparing the performance of neural differential equations (NODEs) against neural state-space models (NSSM) and classical linear subspace identification methods (SID).
In the numerical experiments, we evaluate the methods on eight different dynamical systems described by ordinary differential equations.
We compare the performance on the open-loop prediction accuracy, sensitivity to hyperparameters, and inference computational time. 
   We found that continuous NODEs typically outperform NSSM and linear models by order of magnitude in the prediction accuracy ($\sim O(10^2)$ to $\sim O(10^3)$) while being more robust to hyperparameter selection. Across all systems, we also observed the variance in open-loop MSE for the NODE to be less or equal to the variance of the NSSM and SID. These performance gains, however,  come with a cost of increased inference times associated with NODEs. The relative increase in elapsed time per sample is typically by a factor between $1.1$ to $4.5$, but could be as large as $\approx50$.
   The presented results validate the competitive edge of the continuous-time NODEs against discrete-time models for nonlinear system identification tasks.
   In future work, we plan to to investigate the use of NODEs in differentiable predictive control~\cite{drgona2022learning} applications and in parameter estimation problems in gray-box settings~\cite{Rackauckas2020}.

\section*{ACKNOWLEDGMENT}

Funding for this work was provided by the Pacific Northwest National
Laboratory’s (PNNL) Laboratory Directed Research and Development (LDRD) Program.
PNNL is a multiprogram national laboratory operated by Battelle for the United States Department of Energy under DEAC05-
76RL01830.

%%%%%%%%%%%%%%%%%%%%%%%%%%%%%%%%%%%%%%%%%%%%%%%%%%%%%%%%%%%%%%%%%%%%%%%%%%%%%%%%

\bibliographystyle{IEEEtran}
\bibliography{references}

\end{document}